\documentclass[11pt]{article}
\usepackage[dvips]{graphicx}
\usepackage{textcomp}
\usepackage{amsbsy}
\usepackage{latexsym}
\usepackage[mathscr]{eucal}
\usepackage{amsfonts}
\usepackage{amssymb}
\usepackage[usenames]{color}

\usepackage{fullpage}

\def\rnew{\color{magenta}}

\begin{document}
\bibliographystyle{plain}
\title
{
 Convex optimization on Banach Spaces} 
\author{R.A. DeVore   and V.N. Temlyakov\thanks{%
   This research was supported by the Office of Naval Research Contracts  ONR-N00014-08-1-1113,  ONR 
 N00014-09-1-0107;   the NSF Grants   DMS 0915231 and DMS-1160841.  This research was initiated when the second
 author was a visiting researcher at TAMU}
    }
\maketitle

\hbadness=10000
\vbadness=10000
\newtheorem{lemma}{Lemma}[section]
\newtheorem{prop}[lemma]{Proposition}
\newtheorem{cor}[lemma]{Corollary}
\newtheorem{theorem}[lemma]{Theorem}
\newtheorem{remark}[lemma]{Remark}
\newtheorem{example}[lemma]{Example}
\newtheorem{definition}[lemma]{Definition}
\newtheorem{proper}[lemma]{Properties}
\newtheorem{assumption}[lemma]{Assumption}
%
\def\RR{\rm \hbox{I\kern-.2em\hbox{R}}}
\def\NN{\rm \hbox{I\kern-.2em\hbox{N}}}
\def\ZZ{\rm {{\rm Z}\kern-.28em{\rm Z}}}
\def\CC{\rm \hbox{C\kern -.5em {\raise .32ex \hbox{$\scriptscriptstyle
|$}}\kern
-.22em{\raise .6ex \hbox{$\scriptscriptstyle |$}}\kern .4em}}
\def\vp{\varphi}
\def\<{\langle}
\def\>{\rangle}
\def\t{\tilde}
\def\i{\infty}
\def\e{\varepsilon}
\def\sm{\setminus}
\def\nl{\newline}
\def\o{\overline}
\def\wt{\widetilde}
\def\wh{\widehat}
\def\cT{{\cal T}}
\def\cA{{\cal A}}
\def\cI{{\cal I}}
\def\cV{{\cal V}}
\def\cB{{\cal B}}
\def\cF{{\cal F}}
\def \ff{\varphi}

\def\cR{{\cal R}}
\def\cD{{\cal D}}
\def\cP{{\cal P}}
\def\cJ{{\cal J}}
\def\cM{{\cal M}}
\def\cO{{\cal O}}
\def\Chi{\raise .3ex
\hbox{\large $\chi$}} \def\vp{\varphi}
\def\lsima{\hbox{\kern -.6em\raisebox{-1ex}{$~\stackrel{\textstyle<}{\sim}~$}}\kern -.4em}
\def\lsim{\hbox{\kern -.2em\raisebox{-1ex}{$~\stackrel{\textstyle<}{\sim}~$}}\kern -.2em}
\def\[{\Bigl [}
\def\]{\Bigr ]}
\def\({\Bigl (}
\def\){\Bigr )}
\def\[{\Bigl [}
\def\]{\Bigr ]}
\def\({\Bigl (}
\def\){\Bigr )}
\def\L{\pounds}
\def\pr{{\rm Prob}}
\newcommand{\cs}[1]{{\color{magenta}{#1}}}
\def\ds{\displaystyle}
\def\ev#1{\vec{#1}}     
\newcommand{\lt}{\ell^{2}(\nabla)}
\def\Supp#1{{\rm supp\,}{#1}}
\def\R{\mathbb{R}}
\def\E{\mathbb{E}}
\def\nl{\newline}
\def\T{{\relax\ifmmode I\!\!\hspace{-1pt}T\else$I\!\!\hspace{-1pt}T$\fi}}
\def\N{\mathbb{N}}
\def\Z{\mathbb{Z}}
\def\N{\mathbb{N}}
\def\Zd{\Z^d}
\def\Q{\mathbb{Q}}
\def\C{\mathbb{C}}
\def\Rd{\R^d}
\def\gsim{\mathrel{\raisebox{-4pt}{$\stackrel{\textstyle>}{\sim}$}}}
\def\sime{\raisebox{0ex}{$~\stackrel{\textstyle\sim}{=}~$}}
\def\lsim{\raisebox{-1ex}{$~\stackrel{\textstyle<}{\sim}~$}}
\def\div{\mbox{ div }}
\def\M{M}  \def\NN{N}                  
\def\L{{\ell}}               
\def\Le{{\ell^1}}            
\def\Lz{{\ell^2}}
\def\Let{{\tilde\ell^1}}     
\def\Lzt{{\tilde\ell^2}}
\def\Ltw{\ell^\tau^w(\nabla)}
\def\t#1{\tilde{#1}}
\def\la{\lambda}
\def\La{\Lambda}
\def\ga{\gamma}
\def\BV{{\rm BV}}
\def\Ga{\eta}
\def\al{\alpha}
\def\cZ{{\cal Z}}
\def\cA{{\cal A}}
\def\cU{{\cal U}}
\def\argmin{\mathop{\rm argmin}}
\def\argmax{\mathop{\rm argmax}}
\def\prob{\mathop{\rm prob}}

\def \de{\delta}
\def\cO{{\cal O}}
\def\cA{{\cal A}}
\def\cC{{\cal C}}
\def\C{{\cal C}}
\def\cS{{\cal F}}
\def\bu{{\bf u}}
\def\bz{{\bf z}}
\def\bZ{{\bf Z}}
\def\bI{{\bf I}}
\def\cE{{\cal E}}
\def\cD{{\cal D}}
\def\cG{{\cal G}}
\def\cI{{\cal I}}
\def\cJ{{\cal J}}
\def\cM{{\cal M}}
\def\cN{{\cal N}}
\def\cT{{\cal T}}
\def\cU{{\cal U}}
\def\cV{{\cal V}}
\def\cW{{\cal W}}
\def\cL{{\cal L}}
\def\cB{{\cal B}}
\def\cG{{\cal G}}
\def\cK{{\cal K}}
\def\cS{{\cal S}}
\def\cP{{\cal P}}
\def\cQ{{\cal Q}}
\def\cR{{\cal R}}
\def\cU{{\cal U}}
\def\bL{{\bf L}}
\def\bl{{\bf l}}
\def\bK{{\bf K}}
\def\bC{{\bf C}}
\def\X{X\in\{L,R\}}
\def\ph{{\varphi}}
\def\D{{\Delta}}
\def\H{{\cal H}}
\def\bM{{\bf M}}
\def\bx{{\bf x}}
\def\bj{{\bf j}}
\def\bG{{\bf G}}
\def\bP{{\bf P}}
\def\bW{{\bf W}}
\def\bT{{\bf T}}
\def\bV{{\bf V}}
\def\bv{{\bf v}}
\def\bt{{\bf t}}
\def\bz{{\bf z}}
\def\bw{{\bf w}}
\def \span{{\rm span}}
\def \meas {{\rm meas}}
\def\rhom{{\rho^m}}
\def\diff{\hbox{\tiny $\Delta$}}
\def\EE{{\rm Exp}}
\def\lll{\langle}
\def\argmin{\mathop{\rm argmin}}
\def\argmax{\mathop{\rm argmax}}
\def\dJ{\nabla}
\newcommand{\ba}{{\bf a}}
\newcommand{\bb}{{\bf b}}
\newcommand{\bc}{{\bf c}}
\newcommand{\bd}{{\bf d}}
\newcommand{\bs}{{\bf s}}
\newcommand{\bff}{{\bf f}}
\newcommand{\bp}{{\bf p}}
\newcommand{\bg}{{\bf g}}
\newcommand{\by}{{\bf y}}
\newcommand{\br}{{\bf r}}
\newcommand{\be}{\begin{equation}}
\newcommand{\ee}{\end{equation}}
\newcommand{\bea}{$$ \begin{array}{lll}}
\newcommand{\eea}{\end{array} $$}
\def \Vol{\mathop{\rm  Vol}}
\def \mes{\mathop{\rm mes}}
\def \Prob{\mathop{\rm  Prob}}
\def \exp{\mathop{\rm    exp}}
\def \sign{\mathop{\rm   sign}}
\def \sp{\mathop{\rm   span}}
\def \vphi{{\varphi}}
\def \csp{\overline \mathop{\rm   span}}
\newcommand{\KL}{Karh\'unen-Lo\`eve }
%
\newcommand{\beqn}{\begin{equation}}
\newcommand{\eeqn}{\end{equation}}
\def\beginproof{\noindent{\bf Proof:}~ }
\def\endproof{\hfill\rule{1.5mm}{1.5mm}\\[2mm]}

\newenvironment{Proof}{\noindent{\bf Proof:}\quad}{\endproof}

\renewcommand{\theequation}{\thesection.\arabic{equation}}
\renewcommand{\thefigure}{\thesection.\arabic{figure}}

\makeatletter
\@addtoreset{equation}{section}
\makeatother

\newcommand\abs[1]{\left|#1\right|}
\newcommand\clos{\mathop{\rm clos}\nolimits}
\newcommand\trunc{\mathop{\rm trunc}\nolimits}
\renewcommand\d{d}
\newcommand\dd{d}
\newcommand\diag{\mathop{\rm diag}}
\newcommand\dist{\mathop{\rm dist}}
\newcommand\diam{\mathop{\rm diam}}
\newcommand\cond{\mathop{\rm cond}\nolimits}
\newcommand\eref[1]{{\rm (\ref{#1})}}
\newcommand{\iref}[1]{{\rm (\ref{#1})}}
\newcommand\Hnorm[1]{\norm{#1}_{H^s([0,1])}}
\def\int{\intop\limits}
\renewcommand\labelenumi{(\roman{enumi})}
\newcommand\lnorm[1]{\norm{#1}_{\ell^2(\Z)}}
\newcommand\Lnorm[1]{\norm{#1}_{L_2([0,1])}}
\newcommand\LR{{L_2(\R)}}
\newcommand\LRnorm[1]{\norm{#1}_\LR}
\newcommand\Matrix[2]{\hphantom{#1}_#2#1}
\newcommand\norm[1]{\left\|#1\right\|}
\newcommand\ogauss[1]{\left\lceil#1\right\rceil}
\newcommand{\QED}{\hfill
\raisebox{-2pt}{\rule{5.6pt}{8pt}\rule{4pt}{0pt}}%
  \smallskip\par}
\newcommand\Rscalar[1]{\scalar{#1}_\R}
\newcommand\scalar[1]{\left(#1\right)}
\newcommand\Scalar[1]{\scalar{#1}_{[0,1]}}
\newcommand\Span{\mathop{\rm span}}
\newcommand\supp{\mathop{\rm supp}}
\newcommand\ugauss[1]{\left\lfloor#1\right\rfloor}
\newcommand\with{\, : \,}
\newcommand\Null{{\bf 0}}
\newcommand\bA{{\bf A}}
\newcommand\bB{{\bf B}}
\newcommand\bR{{\bf R}}
\newcommand\bD{{\bf D}}
\newcommand\bE{{\bf E}}
\newcommand\bF{{\bf F}}
\newcommand\bH{{\bf H}}
\newcommand\bU{{\bf U}}
\newcommand\cH{{\cal H}}
\newcommand\sinc{{\rm sinc}}
\def\enorm#1{| \! | \! | #1 | \! | \! |}

\newcommand{\dm}{\frac{d-1}{d}}

\let\bm\bf
\newcommand{\bbeta}{{\mbox{\boldmath$\beta$}}}
\newcommand{\bal}{{\mbox{\boldmath$\alpha$}}}
\newcommand{\bbi}{{\bm i}}

\def\nnew{\color{red}}
\def\mnew{\color{blue}}

\def\PB#1{{\color{black}#1}}
\def\WD#1{{\color{black}#1}}

\def\pb#1#2{{\color{red}#2}}

\newcommand{\dI}{\Delta}

\renewcommand{\Pr}{\mathbb{P}}
\maketitle
\date{}

 \begin{abstract}
{Greedy  algorithms which use only function evaluations are applied to convex optimization in a general Banach space $X$.    Along with algorithms that use exact evaluations, algorithms with approximate evaluations are treated.  A priori  upper bounds for the convergence rate of the proposed algorithms are given.   These bounds depend on the smoothness of the objective function and the sparsity or compressibility (with respect to a given dictionary) of a point in $X$ where the minimum is attained.}
\end{abstract}

\section{Introduction}

Convex optimization is an important and well studied subject of numerical analysis.  The canonical setting for such problems is
to find the minimum of a convex function $E$ over a domain in $\R^d$.  Various numerical algorithms have been developed for minimization problems and a priori bounds for their performance have been proven.  We refer the reader to \cite{BL}, \cite{K},   \cite{Nemir}, \cite{N} for the core results in this area.  

In this paper, we are concerned with the more general setting where $E$ is defined on a domain $D$ in a general  Banach space $X$ with norm $\|\cdot\|=\|\cdot\|_X$.   
Thus, our main interest   is in approximating 
\begin{equation}\label{1.1}
E^*:= \inf_{x\in D} E(x).
\end{equation}
   Problems of this type occur in  many important application domains, such as statistical estimation and learning, optimal control, and shape optimization.    Another important motivation for studying such general problems, even for finite dimensional spaces $X$, is that when the dimension $d$ of $X$ is large, we would like to obtain bounds on the convergence rate of a proposed algorithm that are  independent of this dimension.

Solving \iref{1.1} is an example of a high dimensional problem and is known to suffer the curse of dimensionality without
additional assumptions on $E$ which serve to reduce its dimensionality.  These additional assumptions take the form
of smoothness restrictions on $E$ and assumptions which   imply that the minimum in \eref{1.1} is attained on
a   subset of $D$ with additional structure.  Typical assumptions for the latter involve notions of sparsity or compressibility,
which are by now heavily employed concepts for high dimensional problems.  We will always assume that there is   a point $x^*\in D$ where the minimum $E^*$ is attained, $E(x^*)=E^*$.  We do not assume $x^*$ is unique.  The set $D^*=D^*(E)\subset D$ of all points where the minima is attained is convex.

The algorithms studied in this paper utilize dictionaries $\cD$ of $X$.   
A set of elements  $\cD\subset  X$,  whose closed linear span  coincides with $X$ is called a {\it symmetric dictionary} if $\|g\|:=\|g\|_X=1$, for all  $g\in \cD$, and in addition $g\in\cD$ implies $-g\in \cD$.   The simplest example of a dictionary
is $\cD=\{\pm \varphi_j\}_{j\in\Gamma}$ where $\{\varphi_j\}_{j\in\Gamma}$ is  a  Schauder basis for $X$.   In particular for  $X=\R^d$,  one can take the canonical basis $\{e_j\}_{j=1}^d$.

Given such a dictionary $\cD$, there are several types of domains $D$ that are employed in applications.   Sometimes, these domains are the natural domain of the physical problem.   Other time these are constraints imposed on the minimization problem to ameliorate high dimensionality.  We mention the following three common settings.
\vskip .1in
\noindent
{\bf Sparsity Constraints:}  {\it The set $\Sigma_n(\cD)$ of functions
\be
\label{sparse}
g=\sum_{g\in\Lambda}c_gg,\quad \#(\Lambda)=n,
\ee
is called the set of {\it sparse} functions of order $n$ with respect to the dictionary $\cD$.  One common assumption is to minimize $E$ on the domain $D=\Sigma_n(\cD)$, i.e. to look for an $n$ sparse minimizer of  \eref{1.1}.}

\vskip .1in 
\noindent 
{\bf $\ell_1$ constraints:}
 {\it A more general setting is to minimize $E$ over the closure $A_1(\cD)$ (in $X$)  of the convex hull  of $\cD$.    A slightly more general setting is to minimize $E$ over  one of the sets
 \be 
 \label{LM}
 \cL_M:=\{g\in X:\ g/M\in A_1(\cD)\}.
 \ee
   Sometimes $M$ is allowed to vary as in model
  selection or regularization algorithms from statistics.  This is often referred to as $\ell_1$ minimization.}
  \vskip .1in
  \noindent
  {\bf Unconstrained optimization:}  {\it Imposed constraints, such as sparsity or assuming  $D=A_1(\cD)$, are sometimes artificial and may not reflect 
  the original optimization problem.  We consider therefore the   unconstrained minimization where $D=X$.  We always make the assumption that the minimum of $E$ is actually assumed.  Therefore, there is   a point $x^*\in X$ where
  \be
  \label{minassump}
  E^*=E(x^*).
  \ee
  We do not require that $x^*$ is unique.   Notice that in this case the minimum $E^*$ is
  attained on the set 
  \be
  \label{D0}
  D_0:=\{x\in X: \ E(x)\le E(0)\}.
  \ee
  \vskip .1in
\noindent
In what follows, we refer to minimization over $D_0$ to be the {\it unconstrained minimization problem}.}

A typical greedy optimization algorithm builds   approximations to $E^*$ of the form $E(G_m)$, $m=1,2\dots$ where
the elements $G_m$ are built recursively using the dictionary $\cD$ and typically are in $\Sigma_m(\cD)$.   We will always assume that the initial point $G_0$ is chosen as the $0$ element.  Given that $G_m$ has been defined, one first searches for a direction $\varphi_m\in\cD$ for which $E(G_m+\alpha\varphi_m)$ decreases
significantly as $\alpha$ moves away from zero.  Once, $\varphi_m$ is chosen, then one selects $G_{m+1}=G_m+\alpha_m\varphi_m$
or more generally $G_{m+1}=\alpha_m'G_m+\alpha_m\varphi_m$, using some  recipe for choosing $\alpha_m$ or more generally $\alpha_m,\alpha_m'$.  Algorithms of this type are referred to as {\it greedy algorithms}
and will be the object of study in this paper.

There are different strategies for choosing $\varphi_m$ and $\alpha_m,\alpha_m'$ (see, for instance, \cite{Z}, \cite{SSZ}, \cite{CRPW}, \cite{Cl}, \cite{Ja1}, \cite{DHM}, \cite {JS}, \cite{TRD}, and \cite{Ja2}). 
One possibility to choose $\varphi_m$ is to use
the Fr{\' e}chet derivative $E'(G_{m-1})$ of $E$ to choose a steepest descent direction.  This approach has been amply studied
and various convergence results for steepest descent algorithms have been proven, even  for the general Banach space setting.
We refer the reader to the papers \cite{Z,T1,T2} which are representative of the convergence results known in this case.
The selection of $\alpha_m,\alpha_m'$  is commonly referred to as relaxation and is well studied in numerical analysis, although the
 Banach space setting needs additional attention.

Our interest in the present paper are greedy algorithms that do not utilize $E'$.   They are preferred since $E'$ is not given to us
and therefore, in numerical implementations,  must typically be approximated at any given step of the algorithm.  We will analyze several different algorithms of this type which are distinguished from one another by how $G_{m+1}$ is gotten from $G_m$ both in the selection of $\varphi_m$ and the parameters $\alpha_m,\alpha_m'$.     Our  algorithms are built with ideas similar to  the  analogous, well-studied, greedy algorithms for approximation of a given element $f\in X$.  We refer the reader to \cite{Tbook} for a comprehensive description of greedy approximation algorithms.

In this introduction, we limit ourselves to two of the main algorithms studied in this paper.  The first of these, which we call the  Relaxed $E$-Greedy Algorithm (REGA(co)) was introduced in \cite{Z} under the name sequential greedy approximation.
\vskip .1in
\noindent
{\bf Relaxed $E$-Greedy Algorithm (REGA(co)):} 
We define   $G_0 := 0$.  For $m\ge 1$, assuming $G_{m-1}$ has already been defined, we take  $\varphi_m   \in \cD$   and $0\le \lambda_m \le 1$    such that
$$
E((1-\la_m)G_{m-1} + \la_m\varphi_m) = \inf_{0\le \la\le 1;g\in\cD}E((1-\la)G_{m-1} + \la g)
$$
and define
$$
G_m:=  (1-\la_m)G_{m-1} + \la_m\varphi_m.
$$
We assume that there exist such minimizing $\varphi_m$ and $\lambda_m$.
\vskip .1in
\noindent 
We note that the REGA(co) is a modification of the classical Frank-Wolfe algorithm \cite{FW}.   For convenience, we have assumed the existence of a minimizing $\varphi_m$ and $\lambda_m$.  However,  we also analyze algorithms with only approximate implementation which avoids this assumption.

  Observe  that this algorithm is in a sense built for $A_1(\cD)$ because each $G_m$ is obviously in $A_1(\cD)$.
The next algorithm, called the $E$-Greedy Algorithm with Free Relaxation  (EGAFR(co)),  makes some modifications in the relaxation step that will allow it to be applied to the more general unconstrained minimization problem on $D_0$.
\vskip .1in
\noindent
{\bf $E$-Greedy Algorithm with Free Relaxation  (EGAFR(co)).} 
  We define   $G_0  := 0$. For $m\ge 1$, assuming $G_{m-1}$ has already been defined, we take 
$\varphi_m   \in \cD$,  $\alpha_m, \beta_m\in\R$  satisfying  (assuming existence)
$$
E(\alpha_mG_{m-1} + \beta m\varphi_m) = \inf_{ \alpha,\beta\in\R;g\in\cD}E(\alpha G_{m-1} + \beta g)
$$
and define
$$
G_m:=   \alpha_m G_{m-1} + \beta_m\varphi_m.
$$
\vskip .1in

It is easy to see that each of these algorithms has the following monotonicity
  $$
E(G_0)\ge E(G_1) \ge E(G_2) \ge \cdots.
$$

Our main goal in this paper is to understand what can be said a priori about the convergence rate of a specific greedy optimization algorithm of the above form.  Such results are built on two assumptions: (i) the  smoothness of  $E$, (ii)
assumptions that the minimum is attained at a point $x^*$ satisfying a   constraint such as the sparsity or $\ell_1$ constraint.
In what follows to measure the smoothness of $E$, we introduce the   
  modulus of smoothness  
\begin{equation}\label{2.1}
\rho(E,u):=\rho(E,S,u):=\frac{1}{2}\sup_{x\in S, \|y\|=1}|E(x+uy)+E(x-uy)-2E(x)|,
\end{equation}
of $E$ on any given set $S$.
We say that $E$ is uniformly smooth on $S$ if $\rho(E,S,u)/u\to 0$ as $u\to 0$.

The following theorem for  REGA(co) is a prototype of the results proved in this paper.

 \begin{theorem}\label{T1.2}  Let $\displaystyle{E^*:=\inf_{x\in A_1(\cD)}E(x)}$.
 
 \noindent
{\rm (i)}  If $E$ is uniformly smooth on $ A_1(\cD)$, then   the REGA(co) converges:
\be
\label{1.2conv}
\lim_{m\to \infty} E(G_m) =E^*.
\ee

\noindent
{\rm (ii)} If in addition,   $\rho(E,A_1(\cD), u) \le \gamma u^q$, $1<q\le 2$, then
\be
\label{1.2rate}
  E(G_m)-E^*\le C(q,\gamma)m^{1-q},
  \ee
  with a positive constant $C(q,\gamma)$ which depends only on $q$ and $\gamma$.
 \end{theorem}
 The case $q=2$ of this theorem was proved in \cite{Z}.   We prove this theorem in \S \ref{S:proofs}.

As we have already noted, the EGAFR(co) is designed to solve the unconstrained minimization problem where the domain $D=X$.  The performance of this algorithm will depend not only on the smoothness of $E$ but also on the compressibility of a point $x^*\in D^*$ where $E$ takes its minimum.  To quantify this compressibility, we introduce  
\be
\label{Aepsilon}
A(\epsilon):=A(E,\epsilon):= \inf \{M:  \exists y\in \cL_M \ {\rm such\  that\ } \ E(y)-E^*\le \epsilon\}.
\ee
An equivalent way to quantify this compressibility is the error
\be
\label{errore}
e(E,M):=  \inf_{ y\in \cL_M} E(y)-E^*.
\ee
Notice that the functions $A$ and $e$ are pseudo-inverses of one another.

The following theorem states the convergence properties of the EGAFR(co).
\begin{theorem}\label{T1.3} Let $E$ be   uniformly smooth    on $X$ and let $\displaystyle{E^*:=\inf_{x\in X}E(x)=\inf_{x\in D_0}E(x)}$.   
 
\noindent  
 {\rm (i)} The EGAFR(co) converges:
$$
\lim_{m\to \infty} E(G_m) =\inf_{x\in X}E(x)=\inf_{x\in D_0} E(x)=E^*.
$$

\noindent
{\rm (ii)} If the modulus of smoothness of $E$ satisfies $\rho(E,u)\le \gamma u^q$, $1<q\le 2$, then,
  the EGAFR(co) satisfies
\begin{equation}\label{1.2}
E(G_m)-E^* \le C(E,q,\gamma) \epsilon_m,
\end{equation}
where
\be
\label{epsilonm}
\epsilon_m:= \inf \{\epsilon:\   A(\e)^q m^{1-q}\le \epsilon\} . 
\ee
In particular, if for some $r>0$, we have $e(E,M)\le \tilde \gamma M^{-r}$, $M\ge 1$, then
\begin{equation}\label{1.2}
E(G_m)-E^* \le C(E,q,\gamma,\tilde \gamma, r) m^{\frac{1-q}{1+q/r}}.
\end{equation}

\end{theorem}
We note that the EGAFR(co) is a modification of the Weak Greedy Algorithm with Free Relaxation (WGAFR(co)) studied in \cite{T1}.   Also note that if $x^*\in\cL_M$ then the estimate in Theorem \ref{T1.3} reads
\be
\label{Treads}
E(G_m)-E^* \le C(E,q,\gamma) M^qm^{1-q}.
\end{equation}

 We show in the following section how Theorem \ref{T1.2} and Theorem \ref{T1.3} are easily proven using existing results for greedy algorithms.   We also introduce and analyze another greedy algorithm for convex minimization.

The most important results of the present paper are in Section \ref{S:mods} and are motivated by numerical considerations. Very often we cannot calculate the values of $E$ exactly. Even if we can evaluate $E$ exactly,  we may not be able to find the exact value of, say, the quantity
$$\inf_{0\le \la\le1;g\in\cD} E((1-\la)G_{m-1}+\la g)$$
 in the REGA(co). This motivates us to study in \S \ref{S:mods} various modifications of the above algorithms.   For example, the following algorithm, which is an approximate variant of the REGA(co), was introduced in \cite{Z}. 
\vskip .1in
\noindent
 {\bf Relaxed $E$-Greedy Algorithm with error $\de$ (REGA($\de$)).} Let $\de\in(0,1]$.
We define   $G_0 := 0$. Then, for each $m\ge 1$ we have the following inductive definition:
 We take any $\varphi_m   \in \cD$   and $0\le \lambda_m \le 1$   satisfying   
$$
E((1-\la_m)G_{m-1} + \la_m\varphi_m) \le \inf_{0\le \la\le 1;g\in\cD}E((1-\la)G_{m-1} + \la g) +\de
$$
and  define
$$
G_m:=  (1-\la_m)G_{m-1} + \la_m\varphi_m.
$$
\vskip .1in

In Section \ref{S:mods}, we give modifications of this type to the above algorithms and then prove convergence results for these modifications.   For example, the following  convergence result is proven for the REGA($\de$). 
\begin{theorem}
\label{T1.5}
 Let $E$ be a uniformly smooth on $A_1(\cD)$ convex function with modulus of smoothness $\rho(E,u) \le \gamma u^q$, $1<q\le 2$. Then, for the REGA($\de$) we have  
$$
E(G_m)-E^* \le  C(q,\gamma,E,c) m^{1-q}, \quad m\le \de^{-1/q},
$$
where $\displaystyle{E^*:=\inf_{f\in A_1(\cD)}E(x)}$.
\end{theorem}
In the case $q=2$ Theorem \ref{T1.5} was proved in \cite{Z}. 

In the REGA(co) and the REGA($\de$) we solve the univariate convex optimization problem with respect to $\la$
\begin{equation}\label{1.3}
 \inf_{0\le \la\le 1}E((1-\la)G_{m-1} + \la g)
 \end{equation}
 respectively exactly and with an error $\de$. It is well known (see \cite{Nemir}) that there are fast algorithms to solve problem (\ref{1.3}) approximately. We discuss some of them in \S \ref{S:uni}. 
 
 In the EGAFR(co) and the EGAFR($\de$) (see Section 3 for this algorithm) we solve the convex optimization problem for a function on two variables  
\begin{equation}\label{1.4}
 \inf_{\la,w}E((1-w)G_{m-1} + \la g)
 \end{equation}
 respectively exactly and with an error $\de$. We describe in \S \ref{S:multi}  how univariate optimization algorithms can be used for approximate solution of 
 (\ref{1.4}).

\section {Analysis of greedy algorithms}
\label{S:proofs}
We begin this section by showing how to  prove the results for  REGA(co) and EGAFR(co) stated in the introduction, namely Theorems \ref{T1.2} and \ref {T1.3}.   The proof of convergence results for greedy algorithms typically  is done by establishing a recursive inequality for the error $E(G_n)-E^*$.    To analyze the decay of this sequence of errors,  will need   the following
lemma.

\begin{lemma}
\label{decaylemma}
If a sequence $a_m$, $m\ge 0$, of nonnegative numbers satisfies
\be
\label{recur1}
a_m\le a_{m-1}(1-ca_{m-1}^{p}),\quad m\ge 1,
\ee
with $c>0$ and $p>0$.  Then
\be
\label{recur1}
a_n\le Cn^{-1/p},\quad n\ge 1,
\ee
with the constant $C$ depending only on $p$ and $c$.
\end{lemma}

{\bf Proof:} In the case $p\ge 1$ which is used in this paper this follows from Lemma 2.16 of \cite{Tbook}. In the case $p\ge 1$ Lemma \ref{decaylemma} was often used in greedy approximation in Banach spaces (see \cite{Tbook}, Chapter 6). For the general case $p>0$ see Lemma 4.2 of \cite{NP}). \hfill $\Box$
\vskip .1in

To establish a recursive inequality for the error in REGA(co), we will use  the following lemma about REGA(co).
\begin{lemma}\label{L2.4} Let $E$ be a uniformly smooth convex function with modulus of smoothness $\rho(E,u)$. Then, for any $f\in A_1(\cD)$ and the iterations $G_m$ of the REGA(co), we have
\be
\label{iterREGA}
E(G_m) \le E(G_{m-1} )+ \inf_{0\le\la\le 1}(-\la  (E(G_{m-1} )-E(f))+ 2\rho(E, 2\la)),
\quad m=1,2,\dots .
\ee
\end{lemma}

{\bf Proof:}  A similar result  was proved in  Lemma 3.1 of \cite{T1} for a different greedy algorithm denoted by WRGA(co) in \cite{T1}.  In order to distinguish the two algorithms, we denote by 
  $\bar G_m$ the output of WRGA(co).   The relaxation step in WRGA(co) is exactly the same as in our REGA(co).  However the choice of direction     $\bar\varphi_m$  in  WRGA(co)  was based on a maximal gradient descent.   This means that at each step   the $\bar G_{m-1}$ is also possibly different than our $G_{m-1}$ of  REGA(co).
However, an examination of the proof of Lemma 3.1 shows that it did not matter what $\bar G_{m-1}$ is as long as it is in
$\Sigma_{m-1}(\cD)$.   So Lemma 3.1 holds for our  $G_{m-1}$ and if we let $\tilde G_m$ denote the result of applying WRGA(co) to our $G_{m-1}$, then we have  
\beqn
\label{reduce}
E(G_m)\le E(\tilde G_m)\le E(G_{m-1} )+ \inf_{0\le\la\le 1}(-\la  (E(G_{m-1} )-E(f))+ 2\rho(E, 2\la)).
\ee
Here, the first inequality is because REGA(co) minimizes error over all choices of directions $\varphi$ from the dictionary and all choices of the relaxation parameter and thereby is at least as good as the choice from WRGA(co).   The last inequality is from Lemma 3.1 of \cite{T1}.  Thus, we have proven the lemma. \hfill $\Box$

{\bf Proof of Theorem \ref{T1.2}:}  The proof of this theorem is similar to the proof of Theorem 3.1 and Theorem 3.2 in \cite{T1}.  We illustrate the proof of \eref{1.2rate}.   If we denote by $a_m:=E(G_m)-E^*$, then subtracting $E^*$ from both sides of \eref{iterREGA} gives the recursive inequality
\be
\label{rec}
a_m\le a_{m-1} + \inf_{0\le \lambda\le 1} \{-\lambda a_{m-1}+2\gamma \lambda^q\}.
\ee
If we choose $\lambda$ to satisfy
\be\label{choose}
\lambda a_{m-1}=4\gamma (2\lambda)^q
\ee
provided it is not greater than $1$ and choose $1$ otherwise
and use this value in  \eref{rec}, we obtain in case $\lambda \le 1$
\be
\label{decay}
 a_m\le a_{m-1}(1-ca_{m-1}^{\frac{1}{q-1}}),
\ee
with $c>0$ a constant depending only on $\gamma$ and $q$.
This recursive inequality then gives the decay announced in Theorem \ref{T1.2}  because of    Lemma \ref{decaylemma}. The case $\lambda=1$ can be treated as in the proof of Theorem 3.2 from \cite{T1}. \hfill $\Box$

\vskip .1in
\noindent
{\bf Proof of Theorem \ref{T1.3}:}  This proof is derived from results in \cite{T1} in a  similar way to how we have proved Theorem \ref{T1.2} for REGA(co).  
An algorithm, called WGAFR(co), was introduced in \cite{T1} which differs from EGAFR(co) only in how
each $\varphi_m$ is chosen.   One then uses the analysis in WGAFR(co)

 The above discussed algorithms REGA(co) and EGAFR(co) provide sparse approximate solutions to the corresponding optimization problems. These approximate solutions are sparse with respect to the given dictionary $\cD$ but they are not obtained as an expansion with respect to $\cD$. This means that at each iteration of these algorithms we update all the coefficients of sparse approximants. Sometimes it is important to build an approximant in the form of expansion with respect to $\cD$. The reader can find a discussion of greedy expansions in \cite{Tbook}, Section 6.7.  For comparison with the algorithms we have already introduced, we recall a greedy-type algorithm   for unconstrained optimization which uses only function values and builds sparse approximants in the form of expansion that was introduced and analyzed in \cite{T2}.
Let $\C:=\{c_m\}_{m=1}^\infty$ be a fixed sequence of positive numbers. 

\vskip .1in
\noindent
 {\bf $E$-Greedy Algorithm with coefficients $\C$ (EGA($\C$)).} We define  $G_0:=0$. Then, for each $m\ge 1$ we have the following inductive definition:

(i) Let $\ff_m\in\cD$ be such that (assuming existence)
$$
 E(G_{m-1}+c_m\ff_m)=\inf_{g\in\cD}E(G_{m-1}+c_m g). 
$$

(ii) Then define
$$
G_m:=G_{m-1}+c_m\ff_m .
$$
 \vskip .1in
 \noindent
 In the above definition, we can restrict ourselves to positive numbers because of the symmetry of the dictionary $\cD$.  

For the analysis of this algorithm, we will  
assume that the sets
$$
D_C:=\{x:E(x)\le E(0)+C\}
$$
are bounded for all finite $C$.    
We recall two results for the EGA($\C$) that were proved in  \cite{T2} . 

\begin{theorem}\label{T2.2} Let   $\mu(u)=o(u)$ as $u\to 0$ and let $E$ be a uniformly smooth convex function satisfying
\begin{equation}\label{2.5}
 E(x+uy)-E(x)-u\<E'(x),y\>\le 2\mu(u),  
\end{equation}
for $ x\in D_2,$ $\|y\|=1,$ $ |u|\le 1$ . Assume that the coefficients sequence $\C:=\{c_j\}$, $c_j\in[0,1]$ satisfies the conditions
\begin{equation}\label{2.6}
\sum_{k=1}^\infty c_k =\infty,  
\end{equation}
\begin{equation}\label{2.7}
\sum_{k=1}^\infty \mu(c_k)\le 1.  
\end{equation}
Then, for each dictionary $\cD$,  the EGA($\C$) satisfies
$$
\lim_{m\to\infty}E(G_m)=\inf_{x\in X}E(x)=:E^*.
$$
\end{theorem}

\begin{theorem}\label{T2.3} Let $E$ be a uniformly smooth convex function with modulus of smoothness $\rho(E,u)\le \gamma u^q$, $q\in(1,2]$ on $D_2$. 
We set $s:=\frac{2}{1+q}$ and $\C_s:=\{ck^{-s}\}_{k=1}^\infty$ with $c$ chosen in such a way that $\gamma c^q \sum_{k=1}^\infty k^{-sq} \le 1$. Then the 
  EGA($\C_s$)  converges 
  with the following rate: for any $r\in(0,1-s)$
$$
E(G_m)-\inf_{x\in A_1(\cD)}E(x) \le C(r,q,\gamma)m^{-r}.
$$
\end{theorem}

Let us now turn to a brief comparison of the above algorithms and their known convergence rates. The REGA(co) is designed for solving optimization problems on domains $D\subset A_1(\cD)$ and requires that $D^*\cap A_1(\cD)\neq \emptyset$.
The EGAFR(co) is not limited to the $A_1(\cD)$ but applies for any optimization domain as long as $E$ achieves its minimum on a bounded domain.  As we have noted earlier, if there is a point $D^*\cap A_1(\cD)\neq \emptyset$, then EGAFR(co) provides the same convergence rate ($O(m^{1-q})$) as REGA(co).   Thus, EGAFR(co) is more robust and requires the solution of only a slightly more involved minimization at each iteration.

The advantage of EGA($\C$) is that it solves a simpler minimization problem at each iteration since the relaxation parameters are set in advance.  However, it requires knowledge of the smoothness order  $q$ of $E$ and also gives a poorer rate of convergence than REGA(co) and the EGAFR(co).  

To continue this discussion let us consider the very special case where $X=\ell^d_p$ and the dictionary $\cD$ is finite, say $\cD=\{g_j\}_{j=1}^N$. In such a case,  the  existence of $\ff_m$ in all the above algorithms is easily proven. The EGA($\C$) simply uses $Nm$ function evaluations to make $m$ iterations. 
The REGA(co) solves a one-dimensional optimization problem 
at each iteration for each dictionary element, thus $N$ such problems. We discuss this problem in Section 4 and show that each such problem can be solved with exponential accuracy with respect to the number of evaluations needed from $E$.  

\section{Approximate greedy algorithms for convex optimization}
\label{S:mods}

We turn now to the main topic of this paper which is modifications of the above greedy algorithms to allow
imprecise calculations or  less strenuous choices for  descent directions and relaxation parameters.  We begin with a discussion   of the Weak Relaxed Greedy Algorithm WRGA(co) which was introduced and analyzed in \cite{T1} and which we already referred to in \S \ref{S:proofs} .   The  WRGA(co) uses   the gradient to choose a steepest descent direction   at each iteration.  The interesting aspect of WRGA(co), relative to imprecise calculations,  is that it uses   a weakness parameter   $t_m<1$ to allow some relative error in estimating $\sup_{g\in \cD} \<-E'(G_{m-1}),g - G_{m-1}\>$. Here and below we use a convenient   bracket notation: for a functional $F\in X^*$ and an element $f\in X$ we write $F(f)= \<F,f\>$. 
We concentrate on a modification of the second step of WRGA(co). Very often we cannot calculate values of $E$ exactly. Even in case we can evaluate $E$ exactly we may not be able to find the exact value of the $\inf_{0\le \la\le 1}E((1-\la)G_{m-1} + \la\varphi_m)$. This motivates us to study the following modification of the WRGA(co). 
\vskip .1in
\noindent
{\bf Weak Relaxed Greedy Algorithm with error $\de$ (WRGA($\de$)).} Let $\de\in (0,1]$.
We define   $G_0:= 0$. Then, for each $m\ge 1$ we have the following inductive definition.

(1) $\varphi_m := \varphi^{\de,\tau}_m \in \cD$ is taken any element satisfying
$$
\<-E'(G_{m-1}),\varphi_m - G_{m-1}\> \ge t_m \sup_{g\in \cD} \<-E'(G_{m-1}),g - G_{m-1}\>.
$$

(2) Then $0\le \lambda_m \le 1$ is chosen as any number such that
$$
E((1-\la_m)G_{m-1} + \la_m\varphi_m) \le \inf_{0\le \la\le 1}E((1-\la)G_{m-1} + \la\varphi_m)+\de.
$$
With these choices, we define
$$
G_m:=    (1-\la_m)G_{m-1} + \la_m\varphi_m.
$$
\vskip .1in
 Thus, this algorithm differs from the  REGA($\delta$) given in the introduction, only in the choice of the direction $\varphi_m$
 at each step.  Both of these algorithms are directed at solving the minimization of $E$ over $A_1(\cD)$.  The following theorem analyzes the  WRGA($\delta$).
\begin{theorem}\label{T3.1} Let $E$ be  uniformly smooth on $A_1(\cD)$   whose modulus of smoothness $\rho(E,u)$ satisfies
\be
\label{wrgaqcond}
 \rho(E,u) \le \gamma u^q, \quad 1<q\le 2. 
 \ee
 If $t_k =t$, $k=1,2,\dots,$  then the WRGA($\de$) satisfies
\be
\label{wrgarate}
E(G_m)-E^* \le  C(q,\gamma,t,E) m^{1-q}, \quad m\le \de^{-1/q},
\ee
where $\displaystyle{E^*:=\inf_{f\in A_1(\cD)}E(x)}$.
\end{theorem}

 We develop next some results which will be used to prove this theorem.  Let us first note
  that  when $E$ is Fr{\'e}chet differentiable, the convexity of $E$ implies that for any $x,y$ 
\begin{equation}\label{2.2}
E(y)\ge E(x)+\<E'(x),y-x\>
\end{equation}
or, in other words,
\begin{equation}\label{2.3}
E(x)-E(y) \le \<E'(x),x-y\> = \<-E'(x),y-x\>.
\end{equation} 
The following simple lemma holds.
\begin{lemma}\label{L2.1} Let $E$ be Fr{\'e}chet differentiable convex function. Then the following inequality holds for $x\in S$
\begin{equation}\label{2.4}
0\le E(x+uy)-E(x)-u\<E'(x),y\>\le 2\rho(E,u\|y\|).  
\end{equation}
\end{lemma}

We use these remarks to prove the following.
\begin{lemma}\label{L3.1} Let $E$ be  uniformly smooth on $A_1(\cD)$   with modulus of smoothness $\rho(E,u)$. Then, for any $f\in A_1(\cD)$ we have that the WRGA($\de$) satisfies
$$
E(G_m) \le E(G_{m-1} )+ \inf_{0\le\la\le 1}(-\la t_m (E(G_{m-1} )-E(f))+ 2\rho(E, 2\la))+\de,
\quad m=1,2,\dots 
$$
and therefore
\begin{equation}\label{3.1}
E(G_m)-E^* \le E(G_{m-1} )-E^* + \inf_{0\le\la\le 1}(-\la t_m (E(G_{m-1} )-E^*)+ 2\rho(E, 2\la))+\de,
\quad m=1,2,\dots 
\end{equation}
where $E^*:=\inf_{f\in A_1(\cD)}E(x)$.
\end{lemma}
{\bf Proof:} We have
$$
G_m := (1-\la_m)G_{m-1}+\la_m\varphi_m = G_{m-1}+\la_m(\varphi_m-G_{m-1})
$$
and from the definition of $\lambda_m$,
$$
E(G_m) \le \inf_{0\le \la\le 1}E(G_{m-1}+\la(\varphi_m-G_{m-1}))+\de.
$$
By Lemma \ref{L2.1} we have for any $\la$
$$
E(G_{m-1}+\la (\varphi_m-G_{m-1}))
$$
\begin{equation}\label{3.2}
  \le E(G_{m-1}) - \la\<-E'(G_{m-1}),\varphi_m-G_{m-1}\> + 2 \rho(E,2\la)
\end{equation}
and by step (1) in the definition of the WRGA($\de$) and Lemma 2.2 from \cite{T1} (see also Lemma 6.10, p. 343 of \cite{Tbook}) we get
$$
\<-E'(G_{m-1}),\varphi_m-G_{m-1}\> \ge t_m \sup_{g\in \cD} \<-E'(G_{m-1}),g-G_{m-1}\> =
$$
$$
t_m\sup_{\phi\in A_1(\cD)} \<-E'(G_{m-1}),\phi-G_{m-1}\> \ge t_m   \<-E'(G_{m-1}),f-G_{m-1}\>.
$$
From (\ref{2.3}),    we obtain
$$
\<-E'(G_{m-1}),f-G_{m-1}\> \ge E(G_{m-1})-E(f).
$$
Thus,  
$$
E(G_m) \le \inf_{0\le\la\le1} E(G_{m-1}+ \la(\varphi_m-G_{m-1})) +\de 
$$
\begin{equation}\label{3.3}
\le E(G_{m-1}) + \inf_{0\le\la\le1}(-\la t_m (E(G_{m-1})-E(f)) + 2\rho(E,2\la)+\de,  
\end{equation}
which proves the lemma.
\hfill $\Box$

 Finally, for the proof of Theorem \ref{T3.1}, we will need the following result about sequences.
 
\begin{lemma}\label{L3.2} If a nonnegative sequence $a_0,a_1,\dots,a_N$ satisfies  
\begin{equation}\label{3.5}
a_m\le a_{m-1} +\inf_{0\le \la\le 1}(-\la va_{m-1}+B\la^q)+\de, \quad B>0,\quad \de\in(0,1],
\end{equation}
for $m\le N:=[\de^{-1/q}]$, $q\in (1,2]$, then
\be 
\label{L3.2c}
a_m\le C(q,v,B,a_0) m^{1-q},\quad m\le N. 
\ee
\end{lemma}
{\bf Proof:}
By taking $\lambda=0$,  (\ref{3.5}) implies that 
\begin{equation}\label{3.6}
a_m\le a_{m-1}+\de,\quad m\le N.
\end{equation}
  Therefore,  for all $m\le N$ we have
$$
a_m\le a_0+N\delta \le a_0+1, \quad 0\le m\le N.
$$
Now fix any value of $m\in [0,N]$ and define $\la_1:= \left(\frac{va_{m-1}}{2B}\right)^{\frac{1}{q-1}}$, so that 
\begin{equation}\label{3.7}
 \la_1 v a_{m-1}=  2B\la_1^q .
\end{equation}
If $\la_1\le 1$ then 
$$
\inf_{0\le \la\le 1}(-\la va_{m-1}+ B\la^q)\le -\la_1 va_{m-1}+ B\la_1^q 
$$
$$
=-\frac{1}{2}\la_1va_{m-1} = -C_1(q,v,B)a_{m-1}^p,\quad p:=\frac{q}{q-1}.
$$
If $\la_1> 1$ then for all $\la\le\la_1$ we have $\la v a_{m-1}>  2B\la^q$ and specifying $\la=1$ we get
$$
\inf_{0\le \la\le 1}(-\la va_{m-1}+B\la^q)\le-\frac{1}{2}va_{m-1} 
$$
$$
\le -\frac{1}{2}va_{m-1}^p(a_0+1)^{1-p} = -C_1(q,v,a_0)a_{m-1}^p.
$$
Thus, in any case, setting $C_2:=C_2(q,v,B,a_0):=\min(C_1(q,v,B),C_1(q,v,a_0))$ we obtain
from (\ref{3.5})
\begin{equation}\label{3.8}
a_m\le a_{m-1}- C_2a_{m-1}^p +\de,
\end{equation}
holds for all $0\le m\le N$.

Now to establish \eref{L3.2c}, we  let $n\in [0,N]$ be the smallest integer such that 
\be
\label{smallest}
C_2a_{n-1}^p \le 2\de.
\ee
If there is no such $n$,
we set $n=N$.  In view of \eref{3.8},
we have
\be
\label{have1}
a_m\le a_{m-1}-(C_2/2)a_{m-1}^p, \quad 1\le m\le n.
\ee
If we modify the sequence $a_m$ by defining it to be zero if $m>n$, then this modified sequence satisfies \eref{have1}
for all $m$ and Lemma \ref{decaylemma} gives
\be
\label{gives1}
a_m\le C_3 m^{1-q},\quad 1\le m\le n,
\ee
with $C_3$ depending only on $C_2$ and $p$. 

 If $n=N$, we have finished the proof.
If $n<N$,  then, by (\ref{3.6}), we obtain for $m\in[n,N]$
$$
a_m\le a_{n-1} +(m-n+1)\de \le a_{n-1}+N\delta\le a_{n-1}+ NN^{-q}\le [\frac{2\delta}{C_2}]^{1/p}+C_4N^{1-q},
$$
where we have used the definition of $N$.
Since $\delta^{1/p}\le N^{-q/p}=N^{-q+1}$,  we have 
$$
a_m \le C_5N^{1-q}\le C_5m^{1-q},\quad n\le m\le N,
$$
where $C_5$ depends only on $q,v,B,a_0$.
This completes the proof of the lemma. \hfill $\Box$
\vskip .1in
\noindent
{\bf Proof of Theorem \ref{T3.1}:}   We take
$$
a_n:=E(G_n)-E^* \ge 0.
$$
Then, taking into account that $\rho(E,u)\le \gamma u^q$, we get from Lemma \ref{L3.1}
\begin{equation}\label{3.4}
a_m\le a_{m-1} +\inf_{0\le \la\le 1}(-\la ta_{m-1}+2\gamma(2\la)^q)+\de.
\end{equation}
Applying Lemma \ref{L3.2} with $v=t$, $B=2^{1+q}\gamma$ we complete the proof of Theorem \ref{T3.1}.
\hfill $\Box$

We can establish a similar convergence result for the REGA($\delta$).
\begin{theorem}\label{T3.2} Let $E$ be a uniformly smooth on $A_1(\cD)$ convex function with modulus of smoothness $\rho(E,u) \le \gamma u^q$, $1<q\le 2$. Then, for the REGA($\de$) we have  
$$
E(G_m)-E^* \le  C(q,\gamma,E) m^{1-q}, \quad m\le \de^{-1/q},
$$
where $\displaystyle{E^*:=\inf_{f\in A_1(\cD)}E(x)}$.
\end{theorem}
{\bf Proof:} From  the definition of the REGA($\de$), we have
$$
E(G_m) \le \inf_{0\le \la\le 1;g\in\cD}E((1-\la)G_{m-1} + \la g) +\de.
$$
In the same way that we have proved \eref{iterREGA}, we obtain
\begin{equation}\label{3.11}
E(G_m) \le E(G_{m-1} )+ \inf_{0\le\la\le 1}(-\la  (E(G_{m-1} )-E^*)+ 2\rho(E, 2\la) +\de.
\end{equation}
Inequality (\ref{3.11}) is of the same form as inequality (\ref{3.1}) from Lemma \ref{L3.1}. Thus, repeating the above proof of Theorem \ref{T3.1} we complete the proof of Theorem \ref{T3.2}.
\hfill $\Box$
\vskip .1in

We now introduce and analyze an approximate version of the WGAFR(co). 
\vskip .1in
\noindent
{\bf Weak Greedy Algorithm with Free Relaxation and error $\de$ \ (WGAFR($\de$)).} 
Let $\tau:=\{t_m\}_{m=1}^\infty$, $t_m\in[0,1]$, be a weakness  sequence. We define   $G_0  := 0$. Then for each $m\ge 1$ we have the following inductive definition.

(1) $\varphi_m   \in \cD$ is any element satisfying
\begin{equation}\label{3.12}
\<-E'(G_{m-1}),\varphi_m\> \ge t_m  \sup_{g\in\cD}\<-E'(G_{m-1}),g\>.
\end{equation}

(2) Find $w_m$ and $ \lambda_m$ such that
$$
E((1-w_m)G_{m-1} + \la_m\varphi_m) \le \inf_{ \la,w}E((1-w)G_{m-1} + \la\varphi_m) +\de
$$
and define
$$
G_m:=   (1-w_m)G_{m-1} + \la_m\varphi_m.
$$

\begin{theorem}\label{T3.3} Let $E$ be a uniformly smooth convex function on $X$ with modulus of smoothness $\rho(E,D_1,u)\le \gamma u^q$, $1<q\le 2$ and let   $\displaystyle{E^*:=\inf_{x\in X}E(x)=\inf_{x\in D_0}E(x)}$.     Then,  for the WGAFR($\delta$), we have
\begin{equation}\label{1.2}
E(G_m)-E^* \le C(E,q,\gamma) \epsilon_m,\quad m\le \de^{-1/q}
\end{equation}
where
\be
\label{epsilonm}
\epsilon_m:= \inf \{\epsilon:\   A(\e)^q m^{1-q}\le \epsilon\} . 
\ee
and $A(\epsilon)$ is defined by \eref{Aepsilon}.
\end{theorem}
{\bf Proof:} In the proof of Lemma 4.1 of \cite{T1} we established the inequality
$$
\inf_{ \la\ge 0,w}E((1-w)G_{m-1} + \la\varphi_m) \le E(G_{m-1})
$$
\begin{equation}\label{3.13}
   +\inf_{\la\ge 0}(-\la t_m A(\e)^{-1}(E(G_{m-1})-E^*) +2\rho(E,C_0\la)),\quad C_0=C(D_0),
\end{equation} 
under the assumption that $\ff_m$ satisfies (\ref{3.12}) and $G_{m-1}\in D_0$.

In the case of exact evaluations in the WGAFR(co) we had the monotonicity property $E(G_0)\ge E(G_1)\ge\cdots$ which implied that $G_n\in D_0$ for all $n$. In the case of the WGAFR($\de$) inequality (\ref{3.13}) implies 
\begin{equation}\label{3.14}
E(G_m)\le E(G_{m-1})+\de.
\end{equation}
Therefore, for all $m\le N:=[\de^{-1/q}]$ 
$$
E(G_m)\le E(0)+1,
$$
which implies $G_n\in D_1$ for all $n\le N$. 

Denote
$$
a_n:=E(G_n)-E(f^\e).
$$
Inequality (\ref{3.13}) implies
$$
a_m\le a_{m-1} +\inf_{\la\ge 0}(-\la t A(\e)^{-1}a_{m-1} +2\gamma(C_0\la)^q) +\de.
$$
It is similar to (\ref{3.4}) with the only point that we now cannot guarantee that $a_{m-1}\ge0$.
However, if $n$ is the smallest number from $[1,N]$ such that $a_n<0$ then for $m\in[n,N]$ (\ref{3.14}) 
implies easily $a_m\le Cm^{1-q}$. Thus it is sufficient to assume that $a_n\ge 0$. We apply Lemma 
\ref{L3.2} with $v=t A(\e)^{-1}$, $B=2\gamma C_0^q$ and complete the proof.
\hfill $\Box$

We have discussed above two algorithms the WRGA($\delta$) and the REGA($\delta$). 
Results for the REGA($\delta$) (see Theorem \ref{T3.2}) were derived from the proof of the corresponding results for the WRGA($\delta$) (see Theorem \ref{T3.1}). We now discuss a companion algorithm for the WGAFR($\delta$) that uses only function evaluations. 
\vskip .1in
\noindent
{\bf $E$-Greedy Algorithm with Free Relaxation and error $\delta$  (EGAFR($\delta$)).} 
  We define   $G_0  := 0$. For $m\ge 1$, assuming $G_{m-1}$ has already been defined, we take 
$\varphi_m   \in \cD$  $\alpha_m, \beta_m\in\R$  satisfying 
$$
E(\alpha_mG_{m-1} + \beta m\varphi_m) \le  \inf_{ \alpha,\beta\in\R;g\in\cD}E(\alpha G_{m-1} + \beta g) +\delta
$$
and define
$$
G_m:=   \alpha_m G_{m-1} + \beta_m\varphi_m.
$$
\vskip .1in
In the same way as Theorem \ref{T3.2} was derived from the proof of Theorem \ref{T3.1} one can derive the following theorem from the proof of Theorem \ref{T3.3}. 
\begin{theorem}\label{T3.3E} Let $E$ be a uniformly smooth convex function on $X$ with modulus of smoothness $\rho(E,D_1,u)\le \gamma u^q$, $1<q\le 2$ and let   $\displaystyle{E^*:=\inf_{x\in X}E(x)=\inf_{x\in D_0}E(x)}$.     Then,  for the EGAFR($\delta$), we have
\begin{equation}\label{1.2E}
E(G_m)-E^* \le C(E,q,\gamma) \epsilon_m,\quad m\le \de^{-1/q}
\end{equation}
where
\be
\label{epsilonm}
\epsilon_m:= \inf \{\epsilon:\   A(\e)^q m^{1-q}\le \epsilon\} . 
\ee
and $A(\epsilon)$ is defined by \eref{Aepsilon}.
\end{theorem}

Theorem \ref{T2.3} provides the rate of convergence of the EGA($\C$) where we assume that function evaluations are exact and we can find $\inf_{g\in\cD}$ exactly. However, in practice we very often cannot evaluate functions exactly and (or) cannot find the exact value of the $\inf_{g\in\cD}$. In order to address this issue we modify the EGA($\C$) into the following algorithm EGA($\C,\de$). 

{\bf $E$-Greedy Algorithm with coefficients $\C$ and error $\de$ (EGA($\C,\de$)).} Let $\de\in(0,1]$. We define  $G_0:=0$. Then, for each $m\ge 1$ we have the following inductive definition.

(1) $\ff_m^\de\in\cD$ is such that 
$$
 E(G_{m-1}+c_m\ff_m^\de)\le \inf_{g\in\cD}E(G_{m-1}+c_m g)+\de. 
$$

(2) Let
$$
G_m:=G_{m-1}+c_m\ff_m^\de .
$$

We prove an analog of Theorem \ref{T2.3} for the EGA($\C,\de$). 

\begin{theorem}\label{T3.4} Let $E$ be a uniformly smooth convex function with modulus of smoothness $\rho(E,u)\le \gamma u^q$, $q\in(1,2]$ on $D_3$. 
We set $s:=\frac{2}{1+q}$ and $\C_s:=\{ck^{-s}\}_{k=1}^\infty$ with $c\le 1$ chosen in such a way that $\gamma c^q \sum_{k=1}^\infty k^{-sq} \le 1$. Then the 
  EGA($\C_s,\de$)  provides
  the following rate: for any $r\in(0,1-s)$
$$
E(G_m)-E^* \le C(r,q,\gamma)m^{-r},\qquad m\le \de^{-\frac{1}{1+r}},
$$
where $\displaystyle{E^*:=\inf_{x\in A_1(\cD)}E(x)}$.
\end{theorem}
We first accumulate some results that we will use in the proof of this theorem.  Let $N:=[\de^{-\frac{1}{1+r}}]$, where $[a]$ is the integer part of $a$ and let $G_m$, $m\ge 0$ be the sequence generated by the  EGA($\C_s,\de$) .

\noindent 
{\bf Claim 1:} $G_m\in D_3$, i.e. $E(G_m)\le E(0)+3$, for all $0\le m\le N$.

 To see this, let  $t\in(0,1)$ and $\ff_m$ be such that 
\begin{equation}\label{3.15}
\<-E'(G_{m-1}),\ff_m\>\ge tE_\cD(G_{m-1}),\quad E_\cD(G):=\sup_{g\in\cD}\<-E'(G),g\>.  
\end{equation}
Then
$$
\inf_{g\in\cD}E(G_{m-1}+c_mg)\le E(G_{m-1}+c_m\ff_m).
$$
Thus,   it is sufficient to estimate $E(G_{m-1}+c_m\ff_m)$ with $\ff_m$ satisfying (\ref{3.15}). By (\ref{2.4}) under assumption that $G_{m-1}\in D_3$ we get with $\mu(u):=\gamma u^q$
$$
E(G_{m-1}+c_m\ff_m)\le E(G_{m-1}) +c_m\<E'(G_{m-1}),\ff_m\>+2\mu(c_m).
$$
Using the definition of $\ff_m$,  we obtain
\begin{equation}\label{3.16}
E(G_{m-1}+c_m\ff_m)\le E(G_{m-1})-c_mtE_\cD(G_{m-1})+2\mu(c_m).
\end{equation}
We now prove by induction that $G_m\in D_3$ for all $m\le N$. Indeed, clearly $G_0\in D_3$.  Suppose that $G_k\in D_3$, $k=0,1,\dots, m-1$, then (\ref{3.16}) holds for all $k=1,\dots,m$ instead of $m$ and, therefore,
$$
E(G_m)\le E(0)+2\sum_{k=1}^m\mu(c_k)+m\de \le E(0)+3,
$$
proving the claim. 

We also need the following lemma from \cite{T2}. 
\begin{lemma}\label{L3.3} If  $f\in \cL_A$, then for
$$
G_k:=\sum_{j=1}^kc_j\ff_j,\quad \ff_j\in\cD,\quad j=1,\dots,k,
$$
we have
$$
E_\cD(G_k)\ge(E(G_k)-E(f))/(A+A_k),\quad A_k:=\sum_{j=1}^k|c_j|.
$$
\end{lemma}

\noindent
{\bf Proof of Theorem \ref{T3.4}:}
$E$ attains $E^*$  at a point $x^*\in A_1(\cD)$.  If we start with  (\ref{3.16}) and then use the above lemma with $f=x^*$, fact that  we obtain
\begin{equation}\label{3.17}
E(G_{m} )\le E(G_{m-1})-\frac{tc_m(E(G_{m-1})-E^*)}{1+A_{m-1}}+2\gamma c_m^q+\de. 
\end{equation}
The left hand side of (\ref{3.17}) does not depend on $t$, therefore the inequality holds with $t=1$:
\begin{equation}\label{3.18}
E(G_{m} )\le E(G_{m-1})-\frac{c_m(E(G_{m-1})-E^*)}{1+A_{m-1}}+2\gamma c_m^q+\de. 
\end{equation}
We have
$$
A_{m-1}=c\sum_{k=1}^{m-1}k^{-s} \le c(1+\int_1^mx^{-s}dx)=c(1+(1-s)^{-1}(m^{1-s}-1)))
$$
 and
$$
1+A_{m-1}\le 1+ c(1-s)^{-1}m^{1-s}.
$$
Therefore, for $m\ge C_1$ we have with $v:=(r+1-s)/2$
\begin{equation}\label{3.19}
\frac{c_m}{1+A_{m-1}}\ge \frac{v+1-s}{2(m-1)}.  
\end{equation}
To conclude the proof, we  need the following   technical lemma. This lemma is a more general version of Lemma 2.1 from \cite{T1a} (see also Remark 5.1 in \cite{T7} and Lemma 2.37 on p. 106 of \cite{Tbook}).
\begin{lemma}\label{L3.4}  Let four positive numbers $\alpha < \beta \le 1$, $A$, $U\in \mathbb N$ be given and let a sequence $\{a_n\}_{n=1}^\infty$ have the following properties:  $ a_1<A$ and  we have for all $n\ge 2$
 \begin{equation}\label{3.20}
 a_n\le a_{n-1}+A(n-1)^{-\alpha};  
\end{equation}
 if for some $\nu \ge U$ we have
$$
a_\nu \ge A\nu^{-\alpha}
$$
then
\begin{equation}\label{3.21}
a_{\nu + 1} \le a_\nu (1- \beta/\nu). 
\end{equation}
Then there exists a constant $C=C(\alpha , \beta,A,U )$ such that for all $n=1,2,\dots $ we have
$$
a_n \le C n^{-\alpha} .
$$
 \end{lemma}

We apply this lemma with $a_n:= E(G_n)-E^*$, $n\le N$, $a_n:=0$, $n>N$, $\alpha:=r$, $\beta:=v:=(r+1-s)/2$, $U=C_1$ and $A$ specified later. Let us check the conditions (\ref{3.20}) and (\ref{3.21}) of Lemma \ref{L3.4}. It is sufficient to check these conditions for $m<N$. By the inequality
$$
E(G_m)\le E(G_{m-1})+2\rho(E,c_m) +\de \le E(G_{m-1})+2\gamma c^q m^{-sq} +\de
$$
the condition (\ref{3.20}) holds for $A\ge 2\gamma c^q+1$.  Using $sq\ge 1+r$ we get
\begin{equation}\label{3.22}
c_m^q = c^q m^{-sq}\le c^q m^{-1-r},\qquad \de \le m^{-1-r}. 
\end{equation}
Assume that $a_m\ge Am^{-r}$. Setting $A$ to be big enough to satisfy
$$
\de+2\gamma c_m^q \le \frac{A(1-s-\beta)}{2m^{1+r}}
$$
 we obtain from (\ref{3.18}), (\ref{3.19}), and (\ref{3.22})
$$
a_{m+1}\le a_m(1-\beta/m)
$$
provided $a_m\ge Am^{-r}$. Thus (\ref{3.21}) holds. Applying Lemma \ref{L3.4} we get
$$
a_m\le C(r,q,\gamma)m^{-r}. 
$$
This completes the proof of  Theorem \ref{T3.4}. \hfill $\Box$

\section{Univariate convex optimization}
\label{S:uni}

The relaxation step in each of the above algorithms involves either a univariate or bivariate optimization of a convex function. The univariate optimization problem called {\it line search} is well studied in optimization theory (see \cite{Nemir}). 
The purpose of the remaining two sections of this paper is to show that such problems can be solved efficiently. Results of these two sections are known. We present them here for completeness.

In this section we  consider the class $F$  of
convex on $[0,1]$ functions which belong to Lip $1$ class with constant $1$. 
We are interested in how many function evaluations are needed in order to find for a given $\epsilon>0$ and a given $f\in F$ a point $x^\epsilon\in [0,1]$ such that 
$$
f(x^\epsilon) \le \min_{x\in[0,1]} f(x) +\epsilon\quad?
$$
 We begin with a known upper bound.
 \begin{prop}\label{P4.1} If the algorithm described below is applied to any $f\in F$ and $m\in \N$, then after $3+2m$ function evaluations, it produces a point $x_m\in[0,1]$
such that 
\begin{equation}\label{4.1}
f(x_m) \le \min_{x\in[0,1]} f(x) + 2^{-m}.
\end{equation}
\end{prop}
{\bf Proof:} We begin with three function evaluations $f(0)$, $f(1/2)$, and $f(1)$. Without loss of generality assume that $f(0) \le f(1)$.

\noindent
{\bf Case 1:  
$f(0)\le f(1/2)\le f(1)$}.   It follows  from convexity that  $f(x)\ge f(1/2)$ for all $x\in [1/2,1]$ and  hence we can restrict our 
search for a point of minimum to the interval $[0,1/2]$, in other words we delete interval $(1/2,1]$ from consideration. 

\noindent
{\bf Case 2: $f(1/2)<f(0) $.}  We
make   two more evaluations at the points $1/4$ and $3/4$.   It is impossible that both 
$$
f(1/4)<f(1/2) \quad  {\rm and}\quad f(3/4) < f(1/2).
$$
Therefore, at least one of $f(1/4)$, $f(3/4)$ must be $\ge f(1/2)$.
If $f(1/4) \ge f(1/2)$ and $f(3/4) \ge f(1/2)$ in the same way as above we 
delete intervals $[0,1/4)$ and $(3/4,1]$ and continue our search on $[1/4,3/4]$.
If $f(1/4)<f(1/2)$ then we delete $(1/2,1]$ and if $f(3/4)<f(1/2)$ we delete $[0,1/2)$. 

After one iteration we have added 2 function evaluations and reduced our search for a point of minimum to an interval of length $1/2$ with function values at end points and the middle point known to us.  We continue this process to complete the proof of the proposition.  \hfill $\Box$
\vskip .1in
We next analyze what happens if we do not receive the exact values of $f$ when we query in the above algorithm.  We assume that when we query $f$ at a point $x$, we receive the corrupted value $y(x)$ where $|f(x)-y(x)|\le \delta$ for each $x\in [0,1]$.  We assume that we know $\delta$.
\begin{prop}\label{P5.1} Suppose we make function evaluations with an error $\delta$.  The  algorithm described below applied to $f\in F$ and $m\in \N$ takes  $3+2m$ function evaluations and produces a point $x_m\in[0,1]$
such that 
\begin{equation}\label{5.1}
f(x_m) \le \min_{x\in[0,1]} f(x) + 2^{-m} + (4m+1)\de.
\end{equation}
\end{prop}
{\bf Proof:}   In the argument that follows, we use the following property of convex functions. For any $0\le a<b\le c<d\le 1$ we have 
\begin{equation}\label{5.2}
\frac{f(b)-f(a)}{b-a} \le \frac{f(d)-f(c)}{d-c}.
\end{equation}

As in the proof of Proposition \ref{P4.1} we go by cases. At the first iteration we evaluate our function at $0,1/2,1$. Without loss of generality we assume that $y(0)\le y(1)$. 

{\bf A.} Suppose $y(0)\le y(1/2)$. Then $f(0) \le f(1/2) +2\de$ and by (\ref{5.2}) with $a=0$, $b=1/2$, $c=1/2$, $d=x$, $x\in(1/2,1]$ we obtain
$$
f(x)\ge f(1/2)-2\de, \quad x\in [1/2,1].
$$
Therefore, restricting our search for a minimum to $[0,1/2]$ we make an error of  at most $2\de$. 

{\bf B.} Suppose $y(1/2)<y(0)$. In this case,  we make an additional evaluation of the function at $1/4$. 

{\bf Ba.} Suppose $y(1/4)< y(1/2)-2\de$. Then $f(1/4)<f(1/2)$ and by (\ref{5.2}) we obtain that 
$$
\min_{x\in[1/2,1]} f(x) \ge \min_{x\in[0,1/2]} f(x).
$$
Therefore, we can again restrict our search to the interval $[0,1/2]$.

{\bf Bb.} Suppose $y(1/4)\ge y(1/2)-2\de$. In this case we make an additional evaluation of  the function at $3/4$. If $y(3/4)< y(1/2)-2\de$
then as in {\bf Ba} we can restrict our search to the interval $[1/2,1]$. If $y(3/4)\ge y(1/2)-2\de$ we argue as in the case {\bf A} and 
obtain
$$
\min_{x\in[0,1/4]} f(x) \ge \min_{x\in[1/4,1/2]} f(x) - 4\de,\quad \min_{x\in[3/4,1]} f(x) \ge \min_{x\in[1/2,3/4]} f(x)-4\de.
$$
Therefore, we restrict our search to the interval $[1/4,3/4]$ with an error at most $4\de$. 

At each iteration we  add two evaluations and then find that we can restrict our search to an  interval of half the size of the original  while incurring an additional error at most $4\de$. Finally, the  evaluation of $y$ gives us an error at most $\de$ with that of $f$. 
\hfill $\Box$

We note that convexity of functions from $F$ plays a dominating role in obtaining exponential decay of error in Proposition \ref{P4.1}. For instance,
the following simple known statement holds for the Lip$_11$ class. 

\begin{prop}\label{P4.3} Let $\mathcal A (m)$ denote the class of algorithms (adaptive) which use at most $m$ function evaluations and provide an approximate for the minimum value of a function. Then
$$
  \inf_{A\in\mathcal A(m)}\sup_{f\in  {\rm Lip}_11} |\min_{x\in[0,1]}f(x)-A(f)| = \frac{1}{4m}.
$$
\end{prop} 
{\bf Proof:} The upper bound follows from evaluating $f$ at the midpoints $x_j$  of the intervals $[(j-1)/m,j/m]$, $j=1,\dots,m$ and giving the approximate value $\min_{j}f(x_j)-\frac{1}{4m}$. The lower bound follows from the following observation. For any $m$ points $0\le \xi_1<\xi_2<\dots<\xi_m\le 1$ there are two functions $f_1,f_2\in {\rm Lip}_11$ such that $f_1(\xi_j)=f_2(\xi_j) =0$ for all $j$ and $\min_{x}f_1(x) -\min_xf_2(x) \ge \frac{1}{2m}$. 
\hfill $\Box$

\section {Multivariate convex optimization} 
\label{S:multi}

In this section, we discuss an analog of Proposition \ref{P4.1} for $d$-variate convex functions on $[0,1]^d$.
The $d$-variate algorithm is a coordinate wise application of the algorithm from Proposition \ref{P4.1} with an appropriate $\de$. We begin with a simple lemma.

\begin{lemma}\label{L5.1} Let $f(x)$, $x=(x_1,\dots,x_d)\in[0,1]^d$ be a convex on $[0,1]^d$ function. 
Define $x^d:=(x_1,\dots,x_{d-1})\in [0,1]^{d-1}$ and
$$
f_d(x^d):=\min_{x_d}f(x).
$$
Then $f_d(x^d)$ is a convex function on $[0,1]^{d-1}$. 
\end{lemma}
{\bf Proof:} Let  $u,v \in [0,1]^{d-1}$.  Then, there are two points $w,z \in [0,1]^d$ such that
$$
f_d(u)=f(w),\qquad f_d(v)=f(z)
$$ 
and $u=w^d$, $v=z^d$.   From the  convexity of $f(x)$, we have
\begin{equation}\label{5.3}
f(tw+(1-t)z)\le tf(w)+(1-t)f(z)= tf_d(u)+(1-t)f_d(v),\quad t\in[0,1].
\end{equation}
Clearly,
\begin{equation}\label{5.4}
f_d((tw+(1-t)z)^d)\le f(tw+(1-t)z) ,\quad t\in[0,1].
\end{equation}
Inequalities (\ref{5.3}) and (\ref{5.4}) imply that $f_d(u)$ is convex.
\hfill $\Box$
\begin{prop}\label{P5.2}  The  $d$-variate minimization algorithm given below takes as input any $f\in F$ and $m\in \N$  and produces after $(3+2m)^d$ function evaluations a point $x_m\in[0,1]^d$
such that 
\begin{equation}\label{5.5}
f(x_m) \le \min_{x\in[0,1]} f(x) + 2^{-m}(4m+2)^d.
\end{equation}
\end{prop}
{\bf Proof:}  We construct the algorithm by induction. In the case $d=1$, we use the univariate algorithm  from Proposition \ref{P5.1}. Suppose, we have given the algorithm such that the proposition holds for $d-1$. Then, we write
$$
\min_xf(x) = \min_{x_d} \min_{x^d} f(x)
$$
and observe that by Lemma \ref{L5.1} the function $\displaystyle{g(x_d):=\min_{x^d} f(x)}$ is a convex function.  Next, we apply the algorithm from Proposition \ref{P5.1} with $\de = 2^{-m}(4m+2)^{d-1}$ to the function $g$. By our induction assumption we evaluate $g$ with an error at most $\de$. Thus by Proposition \ref{P5.1} we get an error at most
$$
2^{-m}+(4m+1)\de \le 2^{-m}(4m+2)^d.
$$
The total number of evaluations is $(3+2m)^d$.
This completes the proof.
\hfill $\Box$

\vskip .5in

\noindent
Ronald A. DeVore,  Department of Mathematics, Texas A\& M University, College Station, TX 77843, email: rdevore@math.tamu.edu
\vskip .1in
\noindent
Vladimir Temlyakov, Department of Mathematics, University of South Carolina, Columbia, SC 29208, email: temlyak@math.sc.edu 
\end{document}